\documentclass{article} 

\usepackage{geometry} 
\usepackage{natbib}
\usepackage{hyperref}
\usepackage{url}
\usepackage{amsmath}
\usepackage{color}
\usepackage{graphicx}
\usepackage{caption}
\usepackage{subcaption}
\usepackage{amssymb}
\usepackage{xcolor}
\usepackage{multirow}
\usepackage{booktabs}

\usepackage{authblk}

\usepackage[singlelinecheck=false]{caption}

\usepackage[utf8]{inputenc} 

\title{Correcting Nuisance Variation using Wasserstein Distance}

\author{Gil Tabak}
\author{Minjie Fan}
\author{Samuel J. Yang}
\author{Stephan Hoyer}
\author{Geoff Davis}
\affil{Google Inc.
Mountain View, CA 94043
\authorcr
\texttt{\{tabakg, mjfan, samuely, shoyer, geoffd\}@google.com}}
\begin{document}

\maketitle

\begin{abstract}
Profiling cellular phenotypes from microscopic imaging can provide meaningful biological information resulting from various factors affecting the cells. One motivating application is drug development: morphological cell features can be captured from images, from which similarities between different drug compounds applied at different doses can be quantified. The general approach is to find a function mapping the images to an embedding space of manageable dimensionality whose geometry captures relevant features of the input images. An important known issue for such methods is separating relevant biological signal from nuisance variation. For example, the embedding vectors tend to be more correlated for cells that were cultured and imaged during the same week than for those from different weeks, despite having identical drug compounds applied in both cases. In this case, the particular batch in which a set of experiments were conducted constitutes the domain of the data; an ideal set of image embeddings should contain only the relevant biological information (e.g. drug effects). 
{We develop a general framework for adjusting the image embeddings in order to `forget' domain-specific information while preserving relevant biological information. To achieve this, we minimize a loss function based on distances between marginal distributions (such as the Wasserstein distance) of embeddings across domains for each replicated treatment. For the dataset we present results with, the only replicated treatment happens to be the negative control treatment, for which we do not expect any treatment-induced cell morphology changes. We find that for our transformed embeddings (i) the underlying geometric structure is not only preserved but the embeddings also carry improved biological signal; and (ii) less domain-specific information is present.}
\end{abstract}

\section{Introduction}


In the framework where our approach is applicable,
there are inputs (e.g. images) and a map $\mathcal{F}$  sending the inputs to vectors in a low-dimensional space which summarizes information about the inputs.
 $\mathcal{F}$ could either be engineered using specific image features, or learned (e.g. using deep neural networks). 
We will call these vectors `embeddings' and the space to which they belong the `embedding space'. 
Each input may also have corresponding semantic labels and domains, and for inputs with each label and domain pair,  $\mathcal{F}$ produces some distribution of embeddings. 
Semantically meaningful similarities between pairs of inputs can then be assessed by the distance between their corresponding embeddings, using some chosen distance metric. 
Ideally, the embedding distribution of a group of inputs depends only on their label, but often the domain can influence the embedding distribution as well. 
We wish to find an additional map to adjust the embeddings produced by $\mathcal{F}$ so that the distribution of adjusted embeddings for a given label is independent of the domain, while still preserving semantically meaningful distances between distributions of inputs with different labels.

The map  $\mathcal{F}$ can be used for phenotypic profiling of cells.
In this application, images of biological cells perturbed by one of several possible biological stimuli (e.g. various drug compounds at different doses, some of which may have unknown effects) are mapped to embeddings, which are used to reveal similarities among the applied perturbations.

There are a number of ways to extract embeddings from images of cells.
One class of methods such as that used by \citet{ljosa2013comparison} depend on extracting specifically engineered features. In the recent work by \citet{ando2017improving}, a Deep Metric Network pre-trained on consumer photographic images (not microscopic images of cells) described in~\citet{wang2014learning} was used to generate embedding vectors from cellular images, and it was shown that these embeddings clustered drug compounds by their mechanisms of action (MOA) more effectively. See Fig. \ref{fig:flowchart} for exemplary images of different MOAs.

Currently, one of the most important issues with using image embeddings to discriminate the effects of each treatment (i.e. a particular dose of a drug compound, the `label' in the general problem described above) on morphological cell features is nuisance factors related to slight uncontrollable variation in each biological experiment. Many cell imaging experiments are organized into a number of batches of experiments occurring over time, each of which contains a number of sample plates, each of which contains individual wells in which thousands of cells are grown and treatments are applied.
For this application, the `domain' is an instance of one of these hierarchical levels, and embeddings for cells with a given treatment tend to be closer to each other within the same domain than from different ones.
For example, the experimentalist may apply slightly different concentrations or amounts of a drug compound in two wells in which the same treatment was anticipated. Another example is the location of a particular well within a plate or the order of the plate within a batch, which may influence the rate of evaporation, and hence, the appearance of the cells therein. 
Finally, `batch' effects may result from differences in experimental conditions (temperature, humidity) from week to week; they are various instances of this hierarchical level that we will consider as `domains' in this work.

Many efforts have been made to correct for nuisance variation such as batch effects, especially for microarray gene expression data. 
The simplest method is data normalization such as mean-centering, standardization and quantile normalization. However, data normalization is often not sufficient to ensure the correction of batch effects so that more advanced methods have been developed.
\citet{alter2000singular} proposed a singular value decomposition based method that filters out the eigengenes (and eigenarrays) representing noise or experimental artifacts. 
\citet{benito2004adjustment} used linear discrimination methods such as distance weighted discrimination to adjust for batch biases. \citet{johnson2007adjusting} proposed parametric and non-parametric empirical Bayes frameworks (i.e. ComBat) that remove the additive and multiplicative batch effects.
\citet{leek2007capturing} introduced Surrogate Variable Analysis (SVA) to overcome the problems caused by heterogeneity in expression studies by identifying the effect of the hidden factors that may be the sources of data heterogeneity.
\citet{gagnon2012using} proposed the removal of unwanted variation method (RUV-2) that restricts the factor analysis to negative control genes to infer the unwanted variation.

Most of the aforementioned methods are essentially coordinate-wise in the sense that the batch effect is assumed to affect each dimension (or more specifically each gene for gene expression data) independently.
However, batch effects can be multivariate. This is especially true for the embeddings derived from cellular images. The embedding coordinates are often inter-correlated, and hence can be affected by batch effects jointly.
\citet{lee2014covariance} proposed a multivariate batch adjustment method that can correct for the variance-covariance of data across batches. In particular, they derived an affine transformation that exactly matches the mean vectors and covariance matrices of two batches of data by assuming one of the batches as the target batch (or called the golden batch). The estimation of the transformation was based on a factor model and hard thresholding.

More recent work has started to remove batch effects using deep learning methods. \citet{shaham2017removal} proposed a method that matches the distributions of data in source and target batches using a non-linear transformation based on residual networks, where the distance between the two distributions is measured by the maximum mean discrepancy. This method has two distinct advantages compared to the previous work: it allows for non-linear removal of batch effects and matches the entire distribution instead of the first two moments for the two batches. Matching only the first two moments can be insufficient if the data distribution is multi-modal or highly non-Gaussian. For the embeddings of cellular images, this is likely to happen when individual cells are at different stages of the cell cycle or respond differently to a drug compound and hence form subpopulations. Another class of methods are based on the autoencoder. \citet{amodio2018neuron} learned a latent space of the data using autoencoder, identified batch effect related dimensions, and aligned the distributions of these dimensions across batches. \citet{shaham2018batch} used variational autoencoder to learn a shared encoder to obtain a batch-free encoding of the data, which contains solely biological signal, and also batch dependent decoders that allow for reconstruction of the data to ensure the entire true biological signal will not be lost or distorted. We further discuss these recent methods and how they compare to our method in the conclusion (Section~\ref{sec:future_work}).



In this paper, we address the issue of nuisance variation in image embeddings by transforming the embedding space in a domain-specific way in order to minimize the variation across domains for a given treatment. 
We remark that our main goal is to introduce a general flexible framework to address this problem.
In this framework, we use a metric function measuring the distances among pairs of probability distributions to construct an optimization problem whose solution yields appropriate transformations on each domain.
In our present implementation, the 1-Wasserstein distance is used as a demonstration of a specific choice of the metric that can yield substantial improvements.
In particular, the 1-Wasserstein distance makes few assumptions about the probability distributions of the embedding vectors. Note that Wasserstein distances with higher orders can also be considered.

Our approach is fundamentally different from those which explicitly identify a fixed `target' and `source' distributions given that the target distribution may be difficult to identify.
Instead, we incorporate information from all domains (can be more than two) on an equal footing, transforming all the embeddings to an implicit target distribution. This potentially allows our method to incorporate several replicates of a treatment across different domains to learn the transformations, and not only the negative controls.
We highlight that other distances may be used in our framework, such as the Cramer distance. This may be preferable since the Cramer distance has unbiased sample gradients \citep{bellemare2017cramer}. This could reduce the number of steps required to adjust the Wasserstein distance approximation for each step of training the embedding transformation.
Additionally, we discuss several possible variations and extensions of our method in Section~\ref{sec:future_work}.

\section{Materials and Methods}

\subsection{Problem Description}
Denote the embedding vectors $x_{t,d,p}$ for $t \in T$, $d \in D$, and $p\in I_{t,d}$, where $T$ and $D$ are the treatment and domain labels respectively, and  $I_{t,d}$ is the set of indices for embeddings belonging to treatment $t$ and domain $d$. 
Suppose that $x_{t,d,p}$ are sampled from a probability distribution $\nu_{t,d}$.
Our goal is to `forget' the nuisance variation in the embeddings, which we formalize in the following way.
We wish to find maps $A_d$ transforming the embedding vectors such that the transformed marginal distributions $\tilde \nu_{t,d}$ have the property that for each $t\in T$ and $d_i,d_j\in D$, $\tilde \nu_{t, d_i} \approx \tilde \nu_{t, d_j}$ (for some suitable distance metric between distributions). 
Intuitively, the transformations $A_d$ can be regarded as domain-specific corrections. This is based on the assumption that the embeddings of all the treatments in the same batch are affected by the nuisance variation in the same way.
We do not specify `source' and `target' distributions, and instead transform all the embedding distributions simultaneously. 
The transformations $A_d$ should be small to avoid distorting the underlying geometry of the embedding space, since we do not expect nuisance variation to be very large.
While optimizing the transformations $A_d$, we will at the same time be estimating pairwise distances between transformed embeddings from different domains with the same treatment (which we would like to minimize). This will lead to a minimax problem where we minimize over the transformation parameters and maximize over the distance-estimating parameters.

\subsection{General Approach}
{The 1-Wasserstein distance (hereafter will be simply referred to as the Wasserstein distance) between two probability distributions $\nu_r$ and $\nu_g$  on a compact metric space $\chi$ with metric $\delta$
is given by 
\begin{align}\label{eqn:wd_def}
W(\nu_r, \nu_g) = \inf_{\gamma \in \Pi(\nu_r, \nu_g) } E_{(x,y)\sim \gamma} \delta(x,y).
\end{align}
Here $\Pi(\nu_r,\nu_g)$ is the set of all joint distributions $\gamma(x,y)$ whose marginals are $\nu_r$ and $\nu_g$. This can be intuitively interpreted as the minimal cost of a transportation plan between the probability masses of $\nu_r$ and $\nu_g$.
In our application, the metric space is $\mathbb{R}^n$ and $\delta$ is the Euclidean distance.

Ultimately our goal is to transform pairs of empirical distributions of embeddings so that they become indistinguishable.
We use the Wasserstein distance towards that goal because of two reasons. The first is that when the Wasserstein distance between two distributions $\nu_r$ and $\nu_g$ is zero, they must be identical up to a set of measure zero. Because of this, two empirical distributions drawn from $\nu_r$ and $\nu_g$ cannot be distinguished by any classifier. The second reason is that during the optimization procedure, using the Wasserstein distance yields non-vanishing gradients, which are known to occur for metrics based on the KL-divergence, such as the cross entropy  \citep{arjovsky2017wasserstein}. This is important from a practical point of view because vanishing gradients may halt the solving of the resulting minimax problem in our method. Specifically, we will be maximizing over a set of parameters to obtain estimates for the Wasserstein distance (instead of using a classifier) while at the same time minimizing over the transformation parameters. To highlight the vanishing gradient issue, consider the following case: If we had used a linear classifier in place of the Wasserstein distance, the classifier would become very strongly distinguishing very quickly, having vanishing gradients except near a hyperplane separating the initial empirical distributions, where the gradients become very large.
}

{
Now we seek to extend the usage of the Wasserstein distance to more than two distributions. Given two or more probability distributions, their mean can be defined based on the Wasserstein distance, known as the `Wasserstein barycenter'. Explicitly, the Wasserstein barycenter of $N$ distributions $\nu_1,...,\nu_N$ is defined as the distribution $\mu$ that minimizes
\begin{align}
\frac{1}{N} \sum_{i=1}^N W(\mu, \nu_i).
\end{align}
The Wasserstein barycenter and its computation have been studied in many contexts, such as optimal transport theory \citep{cuturi2014fast, anderes2016discrete}. 
In \citet{tabak2018explanation}, the Wasserstein barycenter has been suggested as a method to remove nuisance variation in high-throughput biological experiments. Two key ingredients of the Wasserstein barycenter are that (i) the nuisance variation is removed in the sense that a number of distinct distributions are transformed into a common distribution, and hence become indistinguishable; and (ii) the distributions are minimally perturbed by the transformations.
}

\begin{figure}[h!]
\begin{center}
\includegraphics[width=0.8\linewidth]{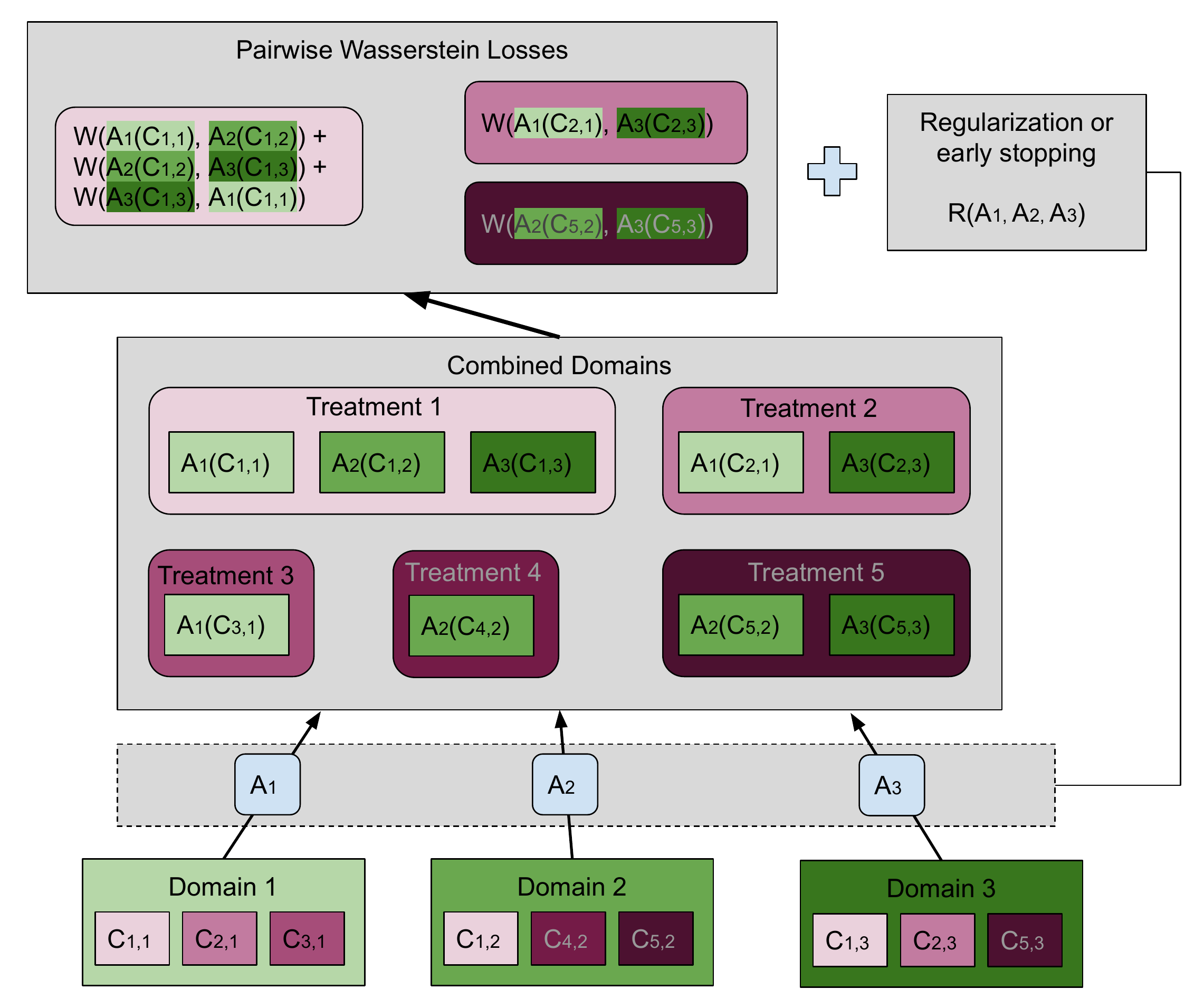}
\end{center}
\caption[.]{
\textbf{Illustration of our approach.} In this case there are three domains (bottom of the plot), each having embeddings for three different treatments. The set of embeddings for treatment $i$ in domain $j$ are represented by $C_{i,j}$.  Each domain has a map $A_j$ transforming all embeddings in domain $j$ to the combined domain space. Treatments are denoted in shades of magenta while domains are in green. Transformed embeddings are collected into groups each having the same treatment. Our goal is to match the distributions within each such group. To achieve this, we wish to minimize the sum of pairwise Wasserstein distances among those distributions that should be matched. A group that contains embeddings from only one domain is not included in the sum. The top box shows the pairwise Wasserstein losses, whose sum is inserted into the loss function. We use an additional neural network to estimate each Wasserstein distance (not shown) since it is not analytically computable. Moreover, there is either a regularization term or early stopping to preserve the geometry of the original embeddings. In this figure, it is illustrated in its most general form as a function of the transformations $A_j$. However, we have found that in our setting it was sufficient to use early stopping, which had essentially the same effect. 
}
\label{fig:model_architecture}
\end{figure}

{
Our method is based on these two requirements, where a separate map is associated with each domain. For each treatment, the average Wasserstein distance among all pairs of transformed distributions across domains is included in the loss function. Specifically, the average Wasserstein distance is formulated as
\begin{align}
\label{equ:our_form}
\frac{2}{N(N-1)}\sum_{i, j =1, i < j}^N W(A_{d_i}(\nu_i), A_{d_j} (\nu_j)),
\end{align}
where the coefficient is the normalizing constant, and $A_{d_i}$ is the map associated with domain $d_i$.
 When multiple treatments are considered, the same number of average Wasserstein distances corresponding to the treatments are included in the loss function.
Thus, (i) is achieved by minimizing a loss function containing pairwise Wasserstein distances. Compared to the ResNet used in  \citet{shaham2017removal}, we achieve (ii) by early stopping to avoid distorting the distributions and hence destroying the biological signal. Adding a regularization term that penalizes the difference between the transformation and the identity map is also feasible.
In Section~\ref{sec:future_work}, we will present another possible formulation that aligns more closely with the idea of the Wasserstein barycenter. The general idea of our method is illustrated in Fig. \ref{fig:model_architecture} by a concrete example.
 }

The Wasserstein distance does not have a closed form except for a few special cases, and must be approximated in some way. 
The Wasserstein distance is closely related to the maximum mean discrepancy (MMD) approximated in  \citet{shaham2017removal} using an empirical estimator based on the kernel method.
This method requires selecting a kernel and relevant parameters.
{In our application, we do not have a fixed `target' distribution, so the kernel parameters would have to be updated during training.}
Instead, we choose to use a method based on the ideas in \citet{arjovsky2017wasserstein} and \citet{gulrajani2017improved} to train a neural network to estimate the Wasserstein distance.
{A similar approach has been proposed in \citet{shen2017adversarial} for domain adaptation}.
To achieve this, we first apply the Kantorovich-Rubinstein duality to (eq.~\ref{eqn:wd_def}):
\begin{align}
\label{equ:wass}
W(\nu_r, \nu_g) = \sup_{\|f\|_L \le1} E_{x\sim \nu_r} \left[f(x)\right] - E_{x\sim \nu_g} \left[f(x)\right].
\end{align}
Here, $\nu_r$ and $\nu_g$ are two probability distributions. The function $f$ is in the space of Lipschitz functions with Lipschitz constant $1$. We will call $f$ the `Wasserstein function' throughout this manuscript.

\subsection{Network Architecture}





Collectively denote the parameters for the transformations $A_d$ by $\Theta_{\text{T}}$. If a particular treatment $t$ is replicated across two or more domains $d_1, d_2,...,d_k$, the Wasserstein distances among the transformed distributions are estimated for all same-treatment domain pairs. Note that the parameters for estimating the Wasserstein distance for each $t$ and pair $d_i,d_j$ are separate. Collectively denote all the Wasserstein estimation parameters by $\Theta_{\text{W}}$. We consider the following loss function
\begin{align}
\label{eqn:raw_loss}
L(\Theta_{\text{T}}, \Theta_{\text{W}})
= \frac{1}{|T|}\sum_{t \in T}
 \frac{2}{|D_t|(|D_t|-1) } \sum_{d_i, d_j \in D_t, i < j}
W(A_{d_i}(\nu_i), A_{d_j} (\nu_j))
+ R(\Theta_{\text{T}}),
\end{align}
where
$$W(A_{d_i}(\nu_i), A_{d_j} (\nu_j))=\sup_{\|f\|_L \le1} E_{x\sim A_{d_i}(\nu_i)} \left[f(x)\right] - E_{x\sim A_{d_j}(\nu_j)} \left[f(x)\right],
$$
and the term inside the $\sup$ operator is called the critic loss.
The function $R(\Theta_{\text{T}})$ is a regularization term for the learned transformations whose purpose is to preserve the geometry of the original embeddings. For example, $R(\Theta_{\text{T}})$ can be proportional to the distance of the transformations from the identity map. Moreover, $D_t$ denotes the domains in which treatment $t$ appears, $T$ denotes all the treatments, and $|\cdot|$ represents the cardinality of a set. In this paper, we ignore $R$ entirely and rely on early stopping, as numerical experiments using regularization have given comparable results. The implementation of early stopping is illustrated in Section \ref{sec:leave-one-compound-out} by leave-one-compound-out cross validation for our particular dataset to preserve the biological information in the data.

%

Each Wasserstein function $f$ should be Lipschitz with Lipschitz constant $1$. For differentiable functions, this is equivalent to the norm of its gradients being bounded by $1$ everywhere.  We use an approach based on \citet{gulrajani2017improved} to impose a soft constraint on the gradient norm.
More specifically, the hard constraint is replaced by a penalty, which is a function of the gradient of the Wasserstein function evaluated at some set of points. The penalty term is weighted by an additional parameter $\gamma$.
Thus, $W(A_{d_i}(\nu_i), A_{d_j} (\nu_j))$ can be approximated by
\begin{align}
\max_{\Theta_{\text{W}}} \left[ W_{t, d_i, d_j} (\Theta_{\text{T}}, \Theta_{\text{W}}) - g_{t, d_i, d_j} (\Theta_{\text{T}}, \Theta_{\text{W}}) \right],
\end{align}
where $W_{t, d_i, d_j}$ is the sample estimate of the critic loss, that is
\begin{align}
\label{equ:wass_approx}
\frac{1}{|I_{t,d_i}|}\sum_{p \in I_{t,d_i}} f_{t,d_i,d_j} (A_{d_i}(x_{t, d_i, p};\Theta_\text{T}); \Theta_\text{W}) 
 - \frac{1}{|I_{t,d_j}|}\sum_{q \in I_{t,d_j}} f_{t,d_i,d_j}(A_{d_j}(x_{t, d_j, q};\Theta_\text{T});\Theta_\text{W}),
\end{align}
and $g_{t,d_i,d_j}$ is the sample estimate of the gradient penalty, that is
\begin{align}
\frac{1}{|J_{t,d_i,d_j}|}
\sum_{z \in J_{t,d_i,d_j}}
H_{t, d_i, d_j}(z; \Theta_{\text{W}}),
\label{equ:gradient_pen_1}
\end{align}
where
\begin{align}
\label{equ:gradient_pen_2}
H_{t, d_i, d_j} (z; \Theta_{\text{W}}) &=
\begin{cases}
\gamma \left(G_{t, d_i, d_j}(z;  \Theta_{\text{W}}) - 1 \right)^2
& \text{if }  G_{t, d_i, d_j}(z; \Theta_{\text{W}}) > 1, \\
0 & \text{otherwise}.
\end{cases} \\
G_{t, d_i, d_j}(z;  \Theta_{\text{W}}) &= 
 \| \nabla_{\Theta_{\text{W}}} f_{t, d_i, d_j}  ( z;\Theta_\text{W} ) \|_2.
 \label{equ:gradient_pen_3}
 \end{align}
 Each Wasserstein function $f_{t, d_i, d_j}$ {in (eq.~\ref{equ:wass_approx})} depends on the parameters $\Theta_\text{W}$, while each transformation $A_d$ depends on the parameters $\Theta_\text{T}$.
For simplicity, we assume that $|I_{t,d_i}| = |I_{t,d_j}|$.
This is a reasonable assumption because in practice, the sets $I_{t,d}$ are chosen as minibatches in stochastic gradient descent.
Since it is impossible to check the gradient everywhere, we use the same strategy as~\citet{gulrajani2017improved}:  choose the intermediate points $\epsilon A_{d_i}(x_{t,d_i,p_k};\Theta_{\text{T}})  + (1 - \epsilon) A_{d_j}(x_{t,d_j,q_k}; \Theta_{\text{T}})$ randomly, where $\epsilon \in U[0,1]$ and $p_k$ and $q_k$ denote the $k$-th elements of $I_{t,d_i}$ and $I_{t,d_j}$, respectively. 
The set of intermediate points are denoted by $J_{t,d_i,d_j}$. 
Intuitively, the reason for sampling along these paths is that the Wasserstein function $f$ whose gradient must be constrained has the interpretation of characterizing the optimal transport between the two probability distributions, and therefore it is most important for the gradient constraint to hold in the intermediate region between the distributions. This is motivated more formally by Proposition 1 in \citet{gulrajani2017improved}, which shows that an optimal transport plan occurs along straight lines with gradient norm 1 connecting
coupled points between the probability distributions. 
 Unlike \citet{gulrajani2017improved}, we impose the gradient penalty only if the gradient norm is greater than $1$. Doing so works better in practice for our application.
We find that the value of $\gamma=10$ used in \citet{gulrajani2017improved} works well in our application, and fix it throughout.
This is an appropriate choice since it is large enough so that the approximation error in the Wasserstein function is small, while not causing numerical difficulties in the optimization routine.

Thus, our objective is to find
\begin{align}
\label{equ:objective}
\hat{\Theta}_\text{T}, \hat{\Theta}_{\text{W}}
 = \text{argmin}_{\Theta_\text{T}} \text{argmax}_{\Theta_\text{W}} L(\Theta_{\text{T}}, \Theta_{\text{W}}).
\end{align}
We use the approach of \citet{ganin2015unsupervised} 
to transform our minimax problem to a minimization problem by adding a `gradient reversal' between the transformed embeddings and the approximated Wasserstein distances. The gradient reversal is the identity in the forward direction, but negates the gradients used for backpropagation. 

\begin{figure}[h!]
\begin{center}
\includegraphics[width=1.0\linewidth]{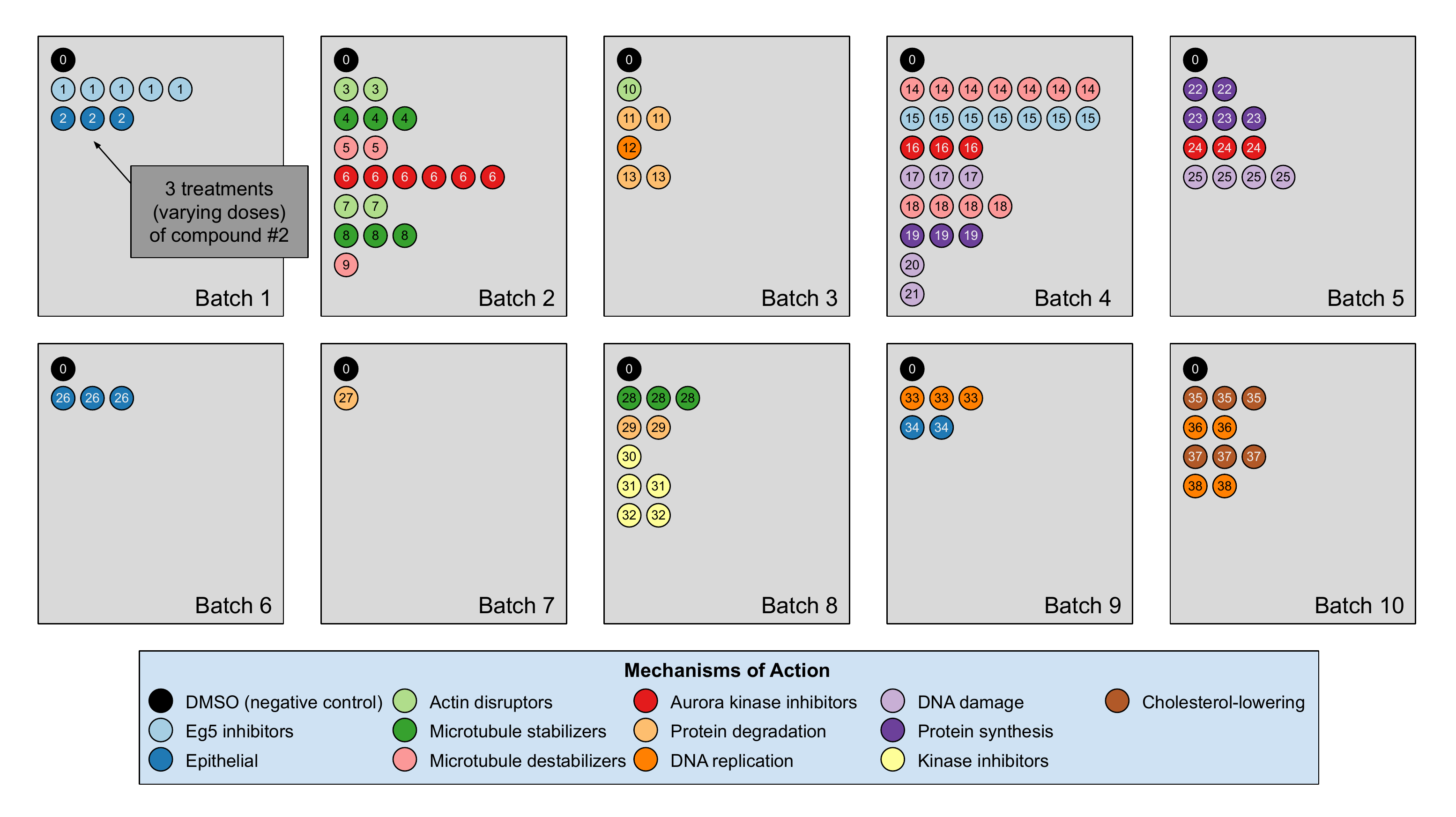}
\end{center}
\caption{\textbf{Illustration of the BBBC021 dataset considered in the paper.} This dataset corresponds to cells prepared across 10 separate batches, to which DMSO (negative control drug compound, labeled by 0) and other drug compounds (labeled by numbers starting from 1) were applied. We use the same subset of treatments (a drug compound with a particular dose) evaluated in  \citet{ljosa2013comparison} and \citet{ando2017improving}. This subset has 103 treatments from 38 drug compounds, each belonging to one of 12 known mechanism of action (MOA) groups (shown in the legend with different colors). Each batch (represented by a rectangle) can contain multiple drug compounds with various doses, each of which is represented by a circle with its label colored according to its MOA. If a drug compound has multiple doses in the same batch, the same number of circles with the exact same pattern are plotted.}\label{fig:data_illustration}
\end{figure}

\subsection{Dataset and Preprocessing Steps}
\label{preprocessing}
{Our analyses are conducted on the image dataset BBBC021 \citep{caie2010high} available from the Broad Bioimage Benchmark Collection \citep{ljosa2012annotated}.} This dataset corresponds to cells prepared on 55 plates across 10 separate batches, and imaged in three color channels (i.e. stains); for a population of negative control cells, a compound (DMSO) with no anticipated drug effect was applied, while various other drug compounds were applied to the remaining cells. We use the same subset of treatments (a drug compound with a particular dose) evaluated in  \citet{ljosa2013comparison} and \citet{ando2017improving}. This subset has 103 treatments from 38 drug compounds, each belonging to one of 12 known mechanism of action (MOA) groups. A diagram illustration of the dataset is shown in Fig. \ref{fig:data_illustration}. It is worth pointing out that only the negative control (i.e. DMSO) has replicates across batches. 

\begin{figure}[h!]
\begin{center}
\includegraphics[width=1.0\linewidth]{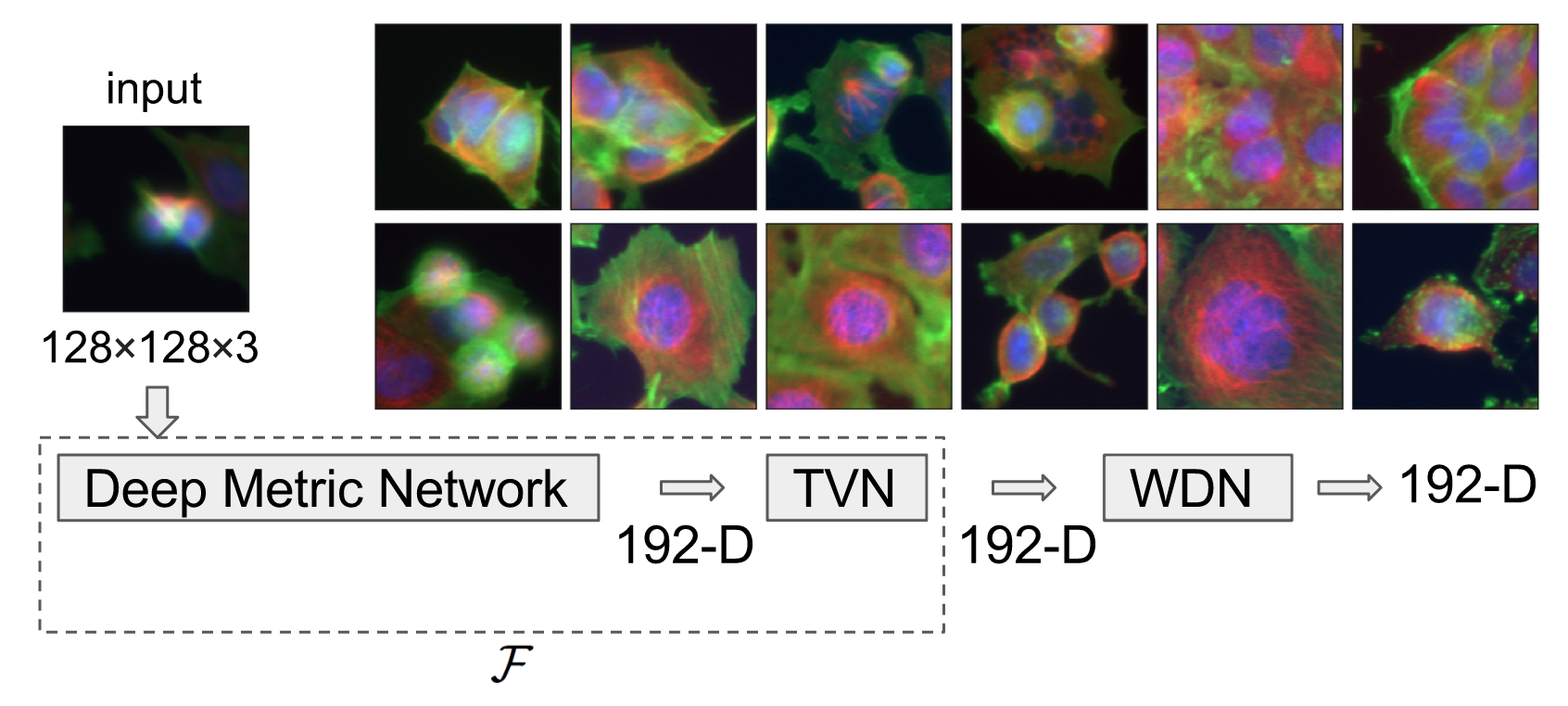}
\end{center}
\caption{\textbf{Flowchart describing the procedure to generate image embeddings using a pre-trained Deep Metric Network and remove nuisance variation from them.} The embedding generator, denoted by $\mathcal{F}$, is described in Section \ref{sec:dnn_emb}, which maps each 128 by 128 color image into a 192-dimensional embedding vector. The nuisance variation removal by our method is denoted by WDN (i.e. Wasserstein Distance Network).
The 12 images on the right show representative images of cells treated with drug compounds with one of the 12 known mechanisms of action (MOA), from the BBBC021 dataset \citep{ljosa2012annotated}.
}
\label{fig:flowchart}
\end{figure}

The treatment is a free label coming from the experiment available for matching treatment replicates across batches (i.e. domains). Multiple drug compounds can share the same MOA, so it is not straightforward to infer the MOA from a drug compound. The MOA information is used for early stopping in model training and model evaluation since it can be used to measure how much MOA-relevant biological information is contained in the embeddings. Sample cell images from the 12 MOA groups are shown in Fig. \ref{fig:flowchart}. 

We apply our method to embeddings that are generated from this dataset in two different ways: hand-engineered features and features extracted by a pre-trained Deep Neural Network.

\subsubsection{Embeddings Based on Hand-engineered Features}

The embeddings are hand-engineered features of length 453 based on specific features for each cell, generated in \citet{ljosa2012annotated}. 
In the original paper, the authors explored a number of ways to pre-process the embeddings.  
We take the approach of factor analysis that gives the best performance. For each coordinate, the 1st percentile of DMSO-treated cells
is set to zero and the 99th percentile is set to 1 for each plate separately. The same
transformations are then applied to all drug compounds on the same plate. After this, factor analysis is applied to the transformed embeddings to reduce the dimensionality to $50$, for which our nuisance variation correction method is conducted.
We will refer to the embeddings generated in this way as \emph{Preprocessed}, and remark this procedure is different from that used for the DNN embeddings presented in Section~\ref{sec:dnn_emb}.
%
%

\subsubsection{Embeddings Based on Deep Neural Network}\label{sec:dnn_emb}

We compute the embeddings for each cell image using the method in \citet{ando2017improving}, summarized as follows. For a 128 by 128 pixel crop around each cell for each of the three color channels, a pre-trained Deep Metric Network generates a $64$-dimensional embedding vector. The three vectors corresponding to the three color channels are concatenated, forming a $192$-dimensional embedding for each cell image. Using the embedding vectors for all cells, a Typical Variation Normalization (TVN) is applied in which the negative controls (i.e. DMSO) are whitened.
Specifically, in the principal component analysis (PCA) basis of only negative control embeddings, an affine transformation is found so that the negative control embeddings have mean zero and identity covariance matrix. The same transformation is then applied to the embeddings of all other cells.
{Note that \citet{ando2017improving} used a different terminology, where TVN includes an additional transformation named CORAL (i.e. correlation alignment), which will be presented and compared to in Section~\ref{sec:comparison}.} 
Our nuisance variation correction method is conducted for the embeddings after TVN.

\subsection{Evaluation Metrics}
\label{sec:metrics}
Our method is evaluated by three metrics, the first two of which measure how much biological signal is preserved in the transformed embeddings, and the last one of which measures how much nuisance variation has been removed.

\subsubsection{k-Nearest Neighbor Mechanism of Action Assignment}
\label{sec:k_nn}

Each drug compound {in} the BBBC021 dataset has a known MOA. A desirable property of embedding vectors is that drug compounds with the same MOA should group closely in the embedding space.
This property can be assessed in the following way using the ground truth MOA label for each treatment.

First, compute the mean $m_X$ of the embeddings for each treatment $X$ in each domain (the negative control is excluded). Find the nearest $k$ neighbors $n_{X,1},n_{X,2},...,n_{X,k}$ of $m_X$ either (i) not belonging to the same compound (abbreviated as NSC) or (ii) not belonging to the same compound or batch (i.e. domain) (abbreviated as NSC NSB), and compute the proportion of them having the same MOA as $m_X$. Our metric is defined as the average of this quantity across all treatment instances $X$ in all domains.
If nuisance variation is corrected by transforming the embeddings, we may expect this metric to increase. 
The reason for excluding same-domain nearest neighbors is to prevent the metric from being interfered by the in-domain correlations.

{
The nearest $k$ neighbors are found based on the cosine distance, which has already been used in existing literature, and in our numerical experiments, we find that it performs better than the Euclidean distance.
Moreover, our k-NN metrics are generalizations of the 1-NN metrics used in \citet{ljosa2013comparison} and \citet{ando2017improving}.
}

{
\subsubsection{Silhouette Score on Mechanism of Action}
\label{sec:sil_index}
Cluster validation measures provide another way of characterizing how well drug compounds with the same MOA group together in the embedding space. In our application, each `cluster' is a chosen MOA containing a group of treatments (the negative control is excluded), and each point in a cluster is the mean of embeddings for a particular treatment (i.e. compound and concentration) and domain.}

{
The Silhouette score is one such measure that compares each point's distance from points in its own cluster to its distance from points in other clusters. Compared to the k-NN metrics as a local metric, Silhouette score is a global metric. It is defined as
\begin{align}
s(i) = \frac{b(i) - a(i)}{\max\{a(i), b(i)\} },
\end{align}
where $a(i)$ is the average distance from point $i$ to all other points in its cluster, and $b(i)$ is the minimum of all average distances from $i$ to all other clusters (i.e. the distance to the closest neighboring cluster) \citep{rousseeuw1987silhouettes}.
The Silhouette score ranges between -1 and 1, with higher values indicating better clustering results. As we did with the k-NN metrics, we also use the cosine distance for the Silhouette score.
}

\subsubsection{Domain Classification Accuracy per Treatment}

\label{sec:domain-classification}
Another metric measures how well domain-specific nuisance information has been `forgotten' (regardless of the biological signal). To achieve this, for each treatment we train a classifier to predict for each embedding the batch (domain) from the set of possible batches (domains) for that treatment.  We evaluate both a linear classifier (i.e. logistic regression) and a random forest with 3-fold cross validation.  If nuisance variation is corrected, the batch (domain) classification accuracy should decrease significantly. Because only the negative control (i.e. DMSO) has replicates across experimental batches in our dataset, we train and evaluate these two batch classifiers on this compound only.

\subsection{Cross Validation and Bootstrapping}

\subsubsection{Early Stopping by Leave-One-Compound-Out Cross Validation}
\label{sec:leave-one-compound-out}

For the model with either early stopping or a regularization term, the hyperparameters (i.e. the stopping time step or the regularization weight) can be selected by a cross validation procedure to avoid overfitting (see \citet{godinez2017multi} for an example).
In particular, we apply this procedure to the case of early stopping.
Each time, an individual drug compound is held out, and the stopping time step is determined by maximizing the average k-NN MOA assignment metric for $k=1,...,4$ on the remaining drug compounds. We can also determine the stopping time step by maximizing the Silhouette score on the remaining drug compounds.

{ For the embeddings transformed at the optimal stopping time step, we evaluate the k-NN MOA assignment metrics for the held-out compound only. The procedure is repeated for all the compounds, and the k-NN MOA assignment metrics are aggregated across all the compounds. 
Intuitively, for each fold of this leave-one-compound-out cross validation procedure, the held-out compound can be regarded as a new compound with unknown MOA, and the hyperparameters are optimized over the compounds with known MOAs.

\subsubsection{Estimating Standard Errors of
the Metrics by Bootstrapping}
\label{sec:standard_error}
To assess whether the improvements in the three evaluation metrics are statistically significant, we estimate the standard errors of the metrics using a nonparametric bootstrap method. Each time, the bootstrap samples are generated by sampling with replacement the embeddings in each well, and the metrics are evaluated using the bootstrap samples. We repeat the procedure for 100 times, and obtain the standard errors (i.e. standard deviation) of the bootstrap estimates of the metrics. 

For the case when using the leave-one-compound-out cross validation discussed in Section~\ref{sec:leave-one-compound-out}, the stopping time was selected separately for each individual bootstrap sample. This way, we are able to evaluate the performance of our method without leaking information about the stopping time from the left-out compound.




\subsection{Model Training}

For simplicity, the embedding transformations are assumed to be affine transformations $A_d(x) = M_d x + b_d$, where $M_d$ is a $d \times d$ matrix and $b_d$ is a $d \times 1$ vector. They are initialized to $M_d = I$, $b_d = 0$, since we wish for the learned transformations to be not too far from the identity map. The intuition is that we can treat nuisance variation as small, random, drug-like perturbations resulting from unobserved covariates. An affine transformation is a first-order approximation to a non-linear perturbation in the sense of the Taylor expansion. It is worth mentioning that we do not expect this assumption to hold in general cases, where nonlinear transformations would be more appropriate.

To approximate each of the Wasserstein functions $f_{t, d_i, d_j}$ in (eq.~\ref{equ:wass_approx}), we use a network consisting of a fully connected layer with Softplus activations followed by a scalar-valued fully connected layer.
The Softplus activation is chosen because the Wasserstein distance estimation it produces is less noisy than other kinds of activations and it avoids the issue of all neurons being deactivated (which can occur for example when using ReLU activations). 

The dimension of the first fully connected layer is set to $2$.  Optimization is done using stochastic gradient descent. For simplicity, the minibatch size for each treatment per iteration step is the same and fixed at $100$ throughout. 
Optimization for both classes of parameters $\Theta_{\text{T}}$ and $\Theta_{\text{W}}$ is done using two separate RMSProp optimizers. 
Prior to training $\Theta_{\text{T}}$, we use a `pre-training' period of $100,000$ time steps to obtain a good initial approximation for the Wasserstein distances. After this, we alternate between training $\Theta_{\text{T}}$ for $50$ time steps and adjusting $\Theta_{\text{W}}$ for a single time step.  

\subsection{Baselines}
\label{sec:comparison}


We compare our approach to the Preprocessed embeddings for the hand-engineered embeddings and the embeddings transformed by TVN for the DNN embeddings, as well as the embeddings transformed by CORAL \citep{sun2016return, ando2017improving} for both types of embeddings. 
CORAL applies a domain-specific affine transformation to the embeddings represented as the rows of a matrix $X_d$ from domain $d$ in the following way. On the negative controls only, the covariance matrix across all domains $C$ as well as the covariance $C_d$ in each domain $d$ are computed.
Then, all embedding coordinates in domain $d$ are aligned by matching the covariance matrices. It is done by computing the aligned embeddings $X^{\text{aligned}}_d = X_d R_d^{-1/2} R^{1/2}$. Here $R_d = C_d + \eta I$ and $R = C + \eta I$ are regularized covariance matrices. Without loss of generality, we specify $C = I$. The regularization weight $\eta$ is set to 1, which is the same as that in~\citet{ando2017improving}.


\section{Results}

Fig. \ref{fig:knn} shows the k-NN MOA assignment accuracy as a function of training time steps for NSC and NSC NSB with $k=1,2,3,4$ using our approach. Knowing when to stop training is a nontrivial problem. We observe that for the DNN embeddings, there is an improvement in all the k-NN metrics to some point, for both filtering based on NSC (Fig. \ref{fig:knn-nsc-deep}) and NSC NSB (Fig. \ref{fig:knn-nsc-nsb-deep}). When continuing to train, these metrics eventually decrease. Meanwhile, the analogous curves using the hand-engineered embeddings (Fig. \ref{fig:knn-nsc} and Fig. \ref{fig:knn-nsc-nsb}) do not improve (is stable at the beginning and then decreases). 

Fig. \ref{fig:violin} shows the distribution of the bootstrap k-NN MOA assignment metrics for NSC and NSC NSB with $k=1,2,3,4$.
We observe that for the DNN embeddings, WDN is at least comparable to CORAL or even sometimes better, and both are generally better than TVN, in preserving MOA-relevant biological information. For the hand-engineered embeddings, however, all perform roughly the same. We also observe that for the DNN embeddings, WDN using the stopping time step based on the Silhouette score outperforms that based on the average k-NN metric. This can be attributed to the Silhouette score being continuous as a function of its inputs, as well as being a global metric.
Tables~\ref{table:nsc} and \ref{table:nsc-nsb} summarize the NSC and NSC NSB k-NN metrics for the hand-engineered embeddings, respectively, while Tables~\ref{table:deep-nsc} and \ref{table:deep-nsc-nsb} summarize the same metrics for the DNN embeddings.
The number in parentheses represents the standard deviation of the bootstrap estimates. 

The biological signal contained in the embeddings can be also captured by the metric the Silhouette score described in Section~\ref{sec:sil_index}. Fig. \ref{fig:silhouette} shows the bootstrap Silhouette scores as a function of training time steps for the hand-engineered and DNN embeddings. For the DNN embeddings, the Silhouette score of WDN significantly increases at the beginning (even above CORAL) and eventually decreases. This suggests that WDN better improves the biological signal than the other methods when the stopping time step is appropriately selected, and over-correction (i.e. being trained for too many time steps) can instead destroy the biological signal. For the hand-engineered embeddings, the Silhouette score only slightly increases at the beginning, reflecting a minor batch effect correction. Tables~\ref{table:sil} and \ref{table:deep-sil} compare the bootstrap Silhouette scores at the time step where the Silhouette score is maximized for the hand-engineered and deep-learned features, respectively. For the hand-engineered embeddings, the Silhouette scores are comparable among the methods, while for the DNN embeddings, the Silhouette score of WDN is significantly better than those of the other methods.

Other than preserving or enhancing the biological signal, we would like to minimize the ability of using embeddings to distinguish which batch a sample comes from given a treatment. For this reason, we use the metric the batch classification accuracy described in Section~\ref{sec:domain-classification}. Fig. \ref{fig:classifications} shows the bootstrap batch classification accuracy using logistic regression and random forest for negative controls as a function of training time steps for the hand-engineered and DNN embeddings. We also include the baseline batch classification accuracy, which is calculated by assigning each embedding coordinate to a Gaussian random variable $\mathcal{N}(0,1)$ independently. In this case, there is no batch information whatsoever, so this represents an effective minimum batch classification accuracy level we should expect. For WDN, all the classification accuracy metrics decrease over training time steps, and eventually become closer to the baseline. 
Tables~\ref{table:sil} and \ref{table:deep-sil} compare the 
bootstrap batch classification accuracy at the time step where the Silhouette score is maximized.
We observe that the batch classification accuracy for WDN is significantly worse than those for TVN (or Preprocessed for the hand-engineered embeddings) and CORAL. All of these suggest the effectiveness of our method in removing the batch effect.

Fig. \ref{fig:PC} compares the first two principal components of embeddings for negative controls across batches among the three methods TVN (or Preprocessed for the hand-engineered embeddings), CORAL and WDN. Each color corresponds to a batch, and there are ten batches in total. These plots are used to show the extent of the batch effect, and to what extent it can be reduced. We observe that for the hand-engineered embeddings (Fig. \ref{fig:PC-hand-eng}), the batch effect is far less severe than that for the DNN embeddings (Fig. \ref{fig:PC-deep}). Furthermore, WDN is better at aligning the distributions of the principal components across batches than the other methods.

\newpage

\begin{figure}[h!]
\begin{center}
\begin{subfigure}{0.5\textwidth}
\includegraphics[width=0.83\linewidth]{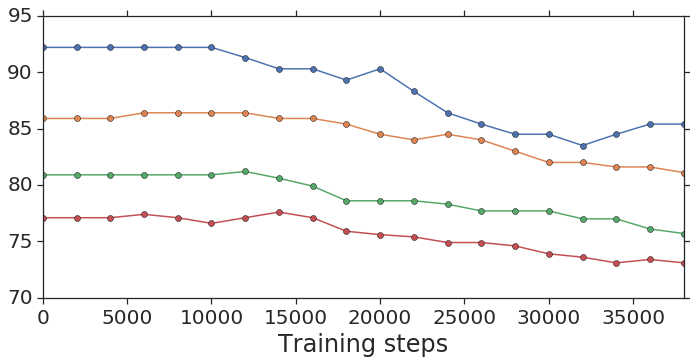}
\caption{NSC accuracy ($\%$) for engineered embeddings.}
\label{fig:knn-nsc}
\end{subfigure}%
\begin{subfigure}{0.5\textwidth}
\includegraphics[width=1.0\linewidth]{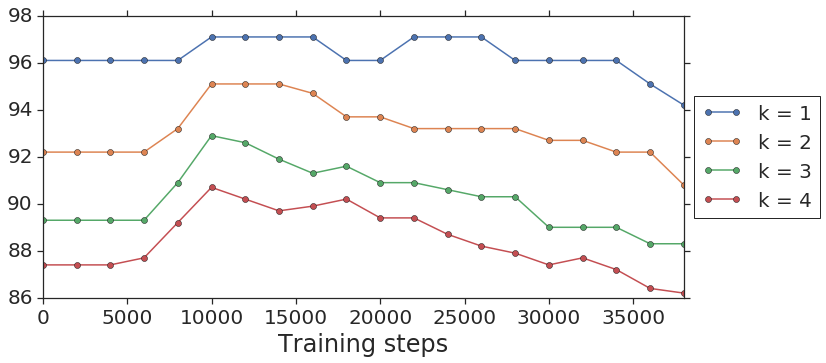}
\caption{NSC accuracy ($\%$) for DNN embeddings.}
\label{fig:knn-nsc-deep}
\end{subfigure}
\begin{subfigure}{0.5\textwidth}
\includegraphics[width=0.83\linewidth]{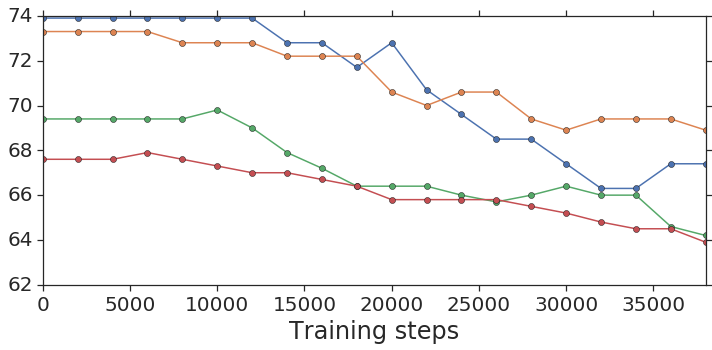}
\caption{NSC NSB accuracy ($\%$) for engineered embeddings.}
\label{fig:knn-nsc-nsb}
\end{subfigure}%
\begin{subfigure}{0.5\textwidth}
\includegraphics[width=0.83\linewidth]{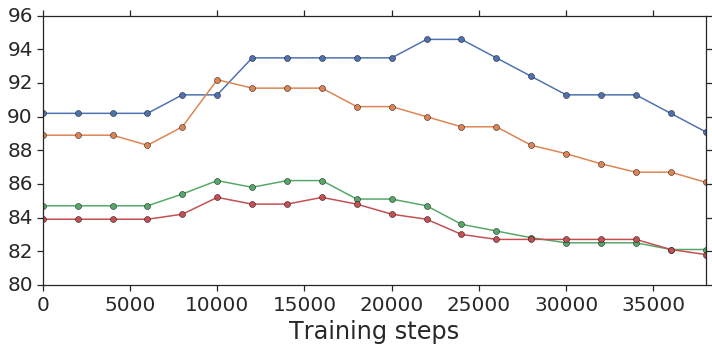}
\caption{NSC NSB accuracy ($\%$) for DNN embeddings.}
\label{fig:knn-nsc-nsb-deep}
\end{subfigure}
\end{center}
\caption{\textbf{Comparison of the k-NN MOA assignment metrics for NSC (i.e. not-same-compound) and NSC NSB (i.e. not-same-compound-or-batch) with $k=1,2,3,4$ over training time steps for the two types of embeddings using our approach (i.e. WDN).} The hand-engineered embedding is shown in (a) and (c), and the DNN embedding is shown in (b) and (d). For the DNN embeddings, there is a significant improvement at some time step over the training. This suggests that the biological signal can be improved if the number of training steps is selected appropriately. For the hand-engineered embeddings, there is little improvement in the metrics.}
\label{fig:knn}
\end{figure}

\begin{figure}[h!]
\begin{center}
\begin{subfigure}{0.5\textwidth}
\includegraphics[width=0.9\linewidth]{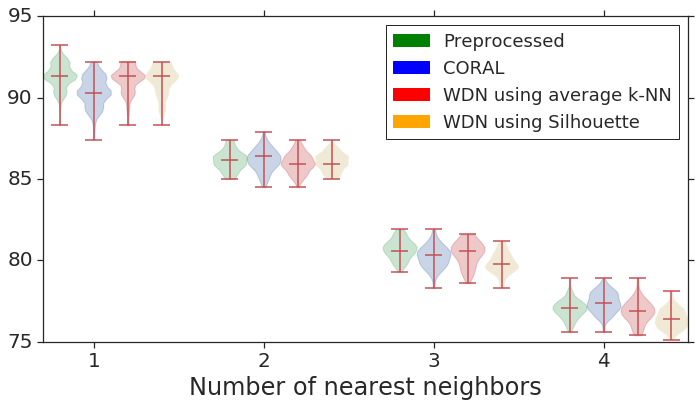}
\caption{NSC accuracy ($\%$) for engineered embeddings.}
\label{fig:violin-nsc}
\end{subfigure}%
\begin{subfigure}{0.5\textwidth}
\includegraphics[width=0.9\linewidth]{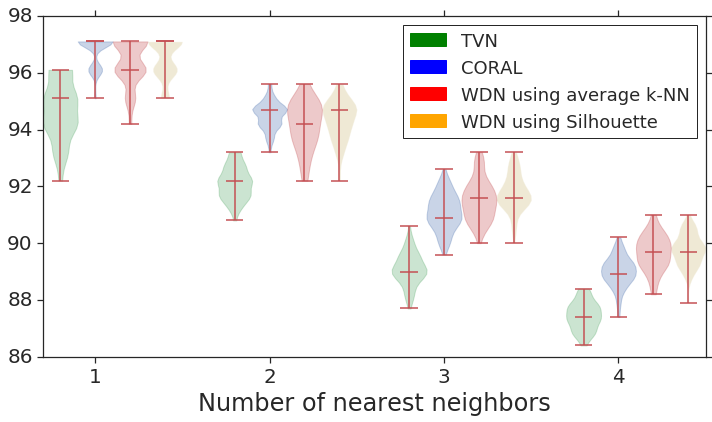}
\caption{NSC accuracy ($\%$) for DNN embeddings.}
\label{fig:violin-nsc-deep}
\end{subfigure}
\begin{subfigure}{0.5\textwidth}
\includegraphics[width=0.9\linewidth]{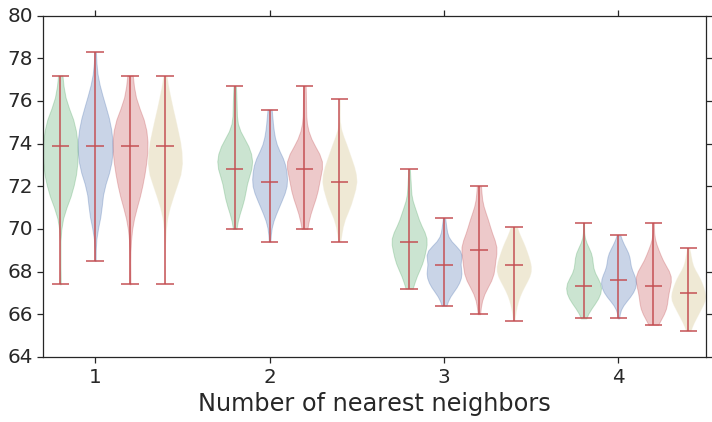}
\caption{NSC NSB accuracy ($\%$) for engineered embeddings.}
\label{fig:violin-nsc-nsb}
\end{subfigure}%
\begin{subfigure}{0.5\textwidth}
\includegraphics[width=0.9\linewidth]{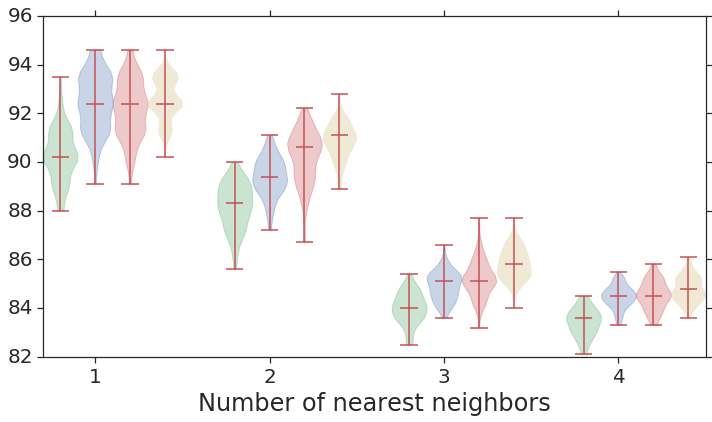}
\caption{NSC NSB accuracy ($\%$) for DNN embeddings.}
\label{fig:violin-nsc-nsb-deep}
\end{subfigure}
\end{center}
\caption{\textbf{Distribution of the bootstrap k-NN MOA assignment metrics for NSC (i.e. not-same-compound) and NSC NSB (i.e. not-same-compound-or-batch) with $k = 1,2,3,4$ for different methods.} We compare TVN (or Preprocessed for the hand-engineered embeddings) with CORAL, and WDN using two different stopping time steps -- based on the average k-NN metric and the Silhouette score. The distribution is shown as a violin plot with three horizontal lines as the 1st, 2nd and 3rd quartiles. For the DNN embeddings, our approach (i.e. WDN) is at least comparable to CORAL or even sometimes better, and both are generally better than TVN, in preserving MOA-relevant biological information. For the hand-engineered embeddings, all perform roughly the same.}\label{fig:violin}
\end{figure}
\newpage

\begin{figure}[h!]
\begin{center}
\begin{subfigure}{0.45\textwidth}
\includegraphics[width=0.9\linewidth]{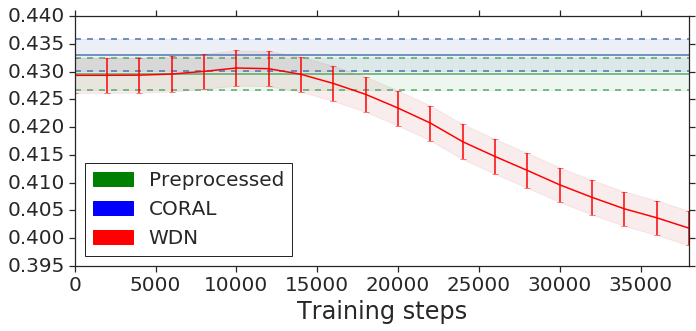}
\caption{Silhouette scores for engineered embeddings over training time steps.}
\end{subfigure}%
\begin{subfigure}{0.45\textwidth}
\includegraphics[width=0.9\linewidth]{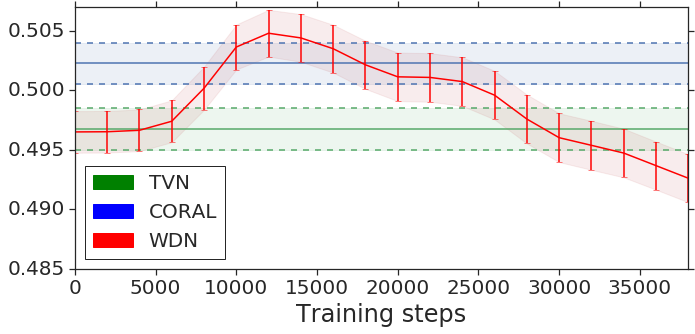}
\caption{Silhouette scores for DNN embeddings over training time steps.}
\end{subfigure}
\end{center}
\caption{\textbf{Bootstrap Silhouette scores on MOA over training time steps for different methods.} We compare the methods TVN (or Preprocessed for the hand-engineered embeddings), CORAL and WDN. Silhouette scores for non-WDN methods are independent of the training time steps. The solid line is the mean of the bootstrap Silhouette scores, and the lower and upper bounds (dashed lines for TVN/Preprocessed and CORAL, and error bars for WDN) are mean minus or plus one standard deviation, respectively. The hand-engineered embedding is shown in (a), and the DNN embedding is shown in (b). For the DNN embeddings, the Silhouette score of WDN significantly increases at the beginning (even above CORAL) and eventually decreases, suggesting MOA-relevant information can be maximized with selection of the model at a particular time step.}
\label{fig:silhouette}
\end{figure}

\begin{figure}[h!]
\begin{center}
\begin{subfigure}{0.45\textwidth}
\includegraphics[width=0.9\linewidth]{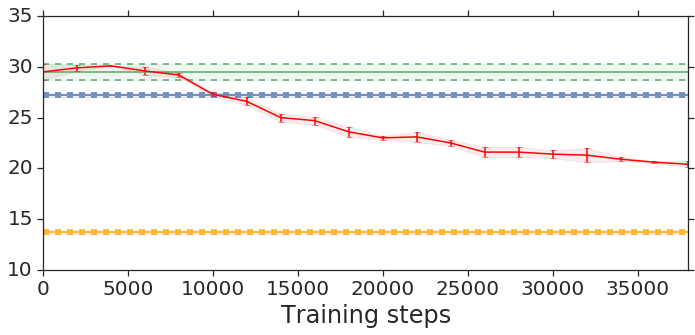}
\caption{Classification accuracy ($\%$) for engineered embeddings using logistic regression.}
\end{subfigure}%
\begin{subfigure}{0.45\textwidth}
\includegraphics[width=0.9\linewidth]{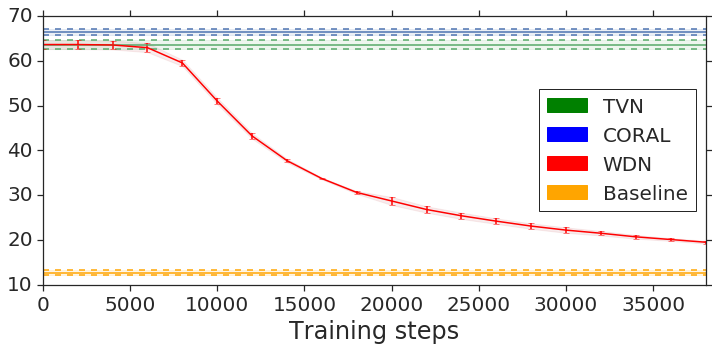}
\caption{Classification accuracy ($\%$) for DNN embeddings using logistic regression.}
\end{subfigure}
\begin{subfigure}{0.45\textwidth}
\includegraphics[width=0.9\linewidth]{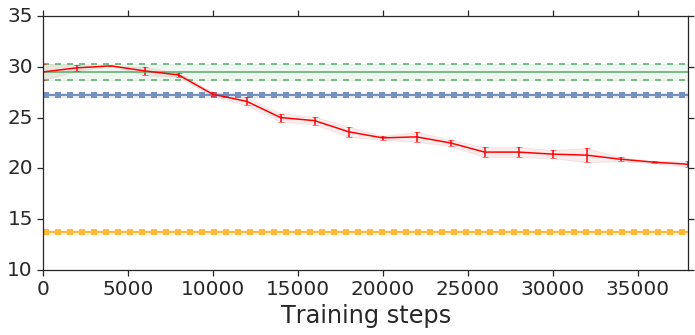}
\caption{Classification accuracy ($\%$) for engineered embeddings using random forest.}
\end{subfigure}%
\begin{subfigure}{0.45\textwidth}
\includegraphics[width=0.9\linewidth]{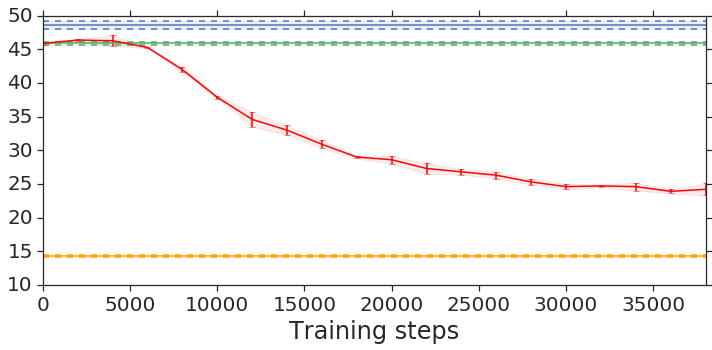}
\caption{Classification accuracy ($\%$) for DNN embeddings using random forest.}
\end{subfigure}
\end{center}
\caption{\textbf{Bootstrap batch (i.e. domain) classification accuracy using logistic regression and random forest for negative controls over training time steps for different methods.} We compare TVN (or Preprocessed for the hand-engineered embeddings), CORAL and WDN. Batch classification accuracy for non-WDN methods are independent of the training time steps. The solid line is the mean of the bootstrap batch classification accuracy, and the lower and upper bounds (dashed lines for TVN/Preprocessed and CORAL, and error bars for WDN) are mean minus or plus one standard deviation, respectively. We also include the baseline batch classification accuracy, which is calculated by assigning each embedding coordinate to a Gaussian random variable $\mathcal{N}(0, 1)$ independently. In this case, there is no batch information whatsoever, so this represents an effective minimum batch classification accuracy level we should expect. The hand-engineered embedding is shown in (a) and (c), and the DNN embedding is shown in (b) and (d). All the classification accuracy metrics for WDN decrease over training time steps, and eventually become closer to chance (i.e. baseline), suggesting WDN successfully removes domain-relevant (i.e. batch) information.}
\label{fig:classifications}
\end{figure}

\newpage

\begin{figure}[h!]
\begin{center}
\begin{subfigure}{0.9\textwidth}
\includegraphics[width=0.9\linewidth]{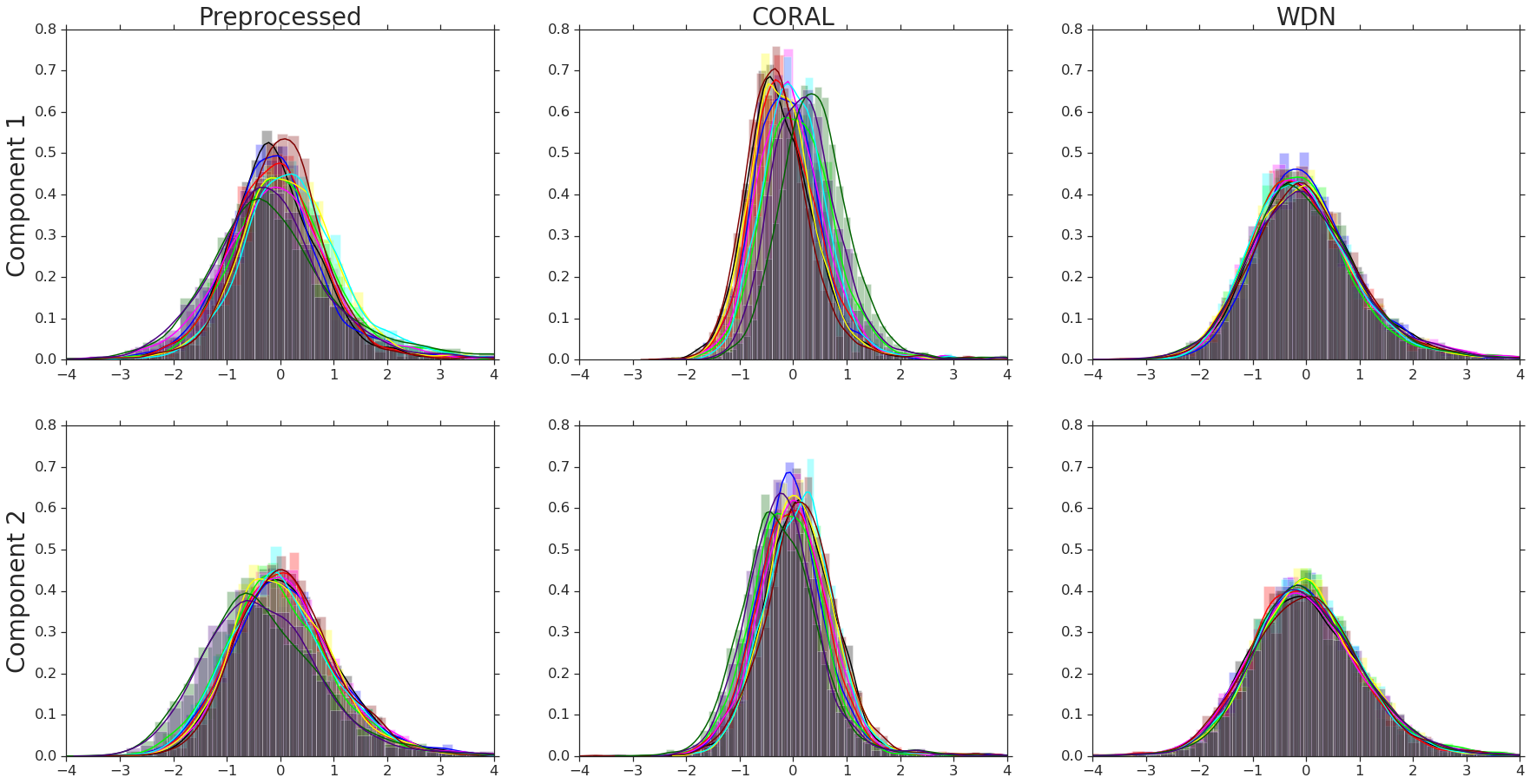}
\caption{The first two principal components of engineered embeddings.}
\label{fig:PC-hand-eng}
\end{subfigure}
\begin{subfigure}{0.9\textwidth}
\includegraphics[width=0.9\linewidth]{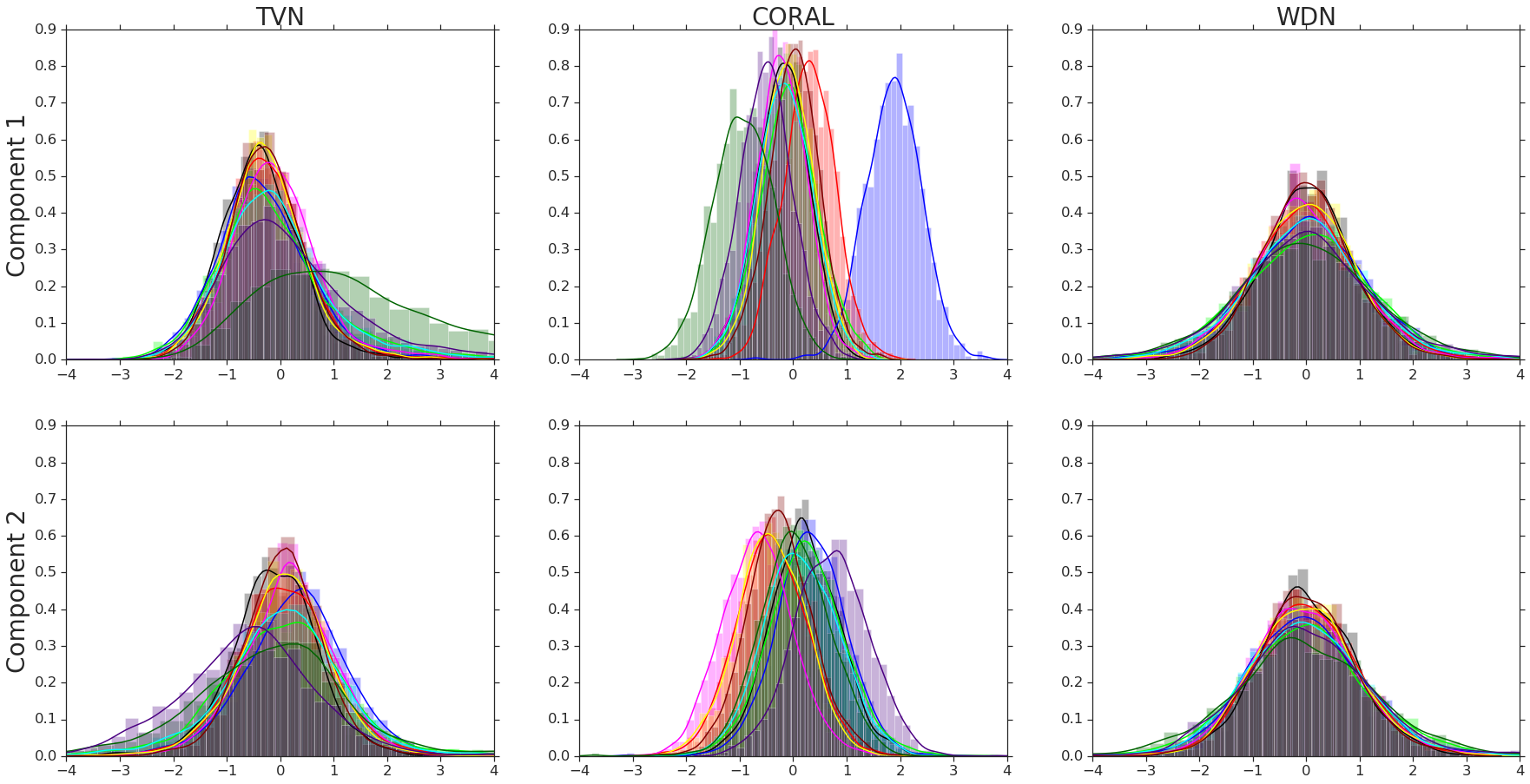}
\caption{The first two principal components of DNN embeddings.}
\label{fig:PC-deep}
\end{subfigure}
\begin{subfigure}{0.75\textwidth}
\includegraphics[width=1.0\linewidth]{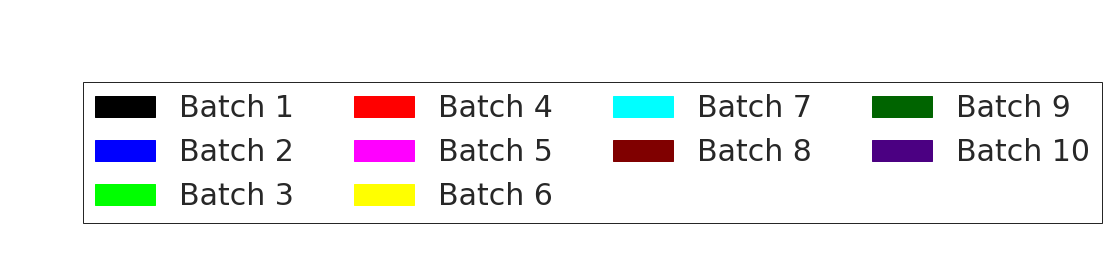}
\end{subfigure}
\end{center}
\caption{\textbf{Comparison of the first two principal components of the two types of embeddings for negative controls for different methods.} We compare the methods TVN (or Preprocessed for the hand-engineered embeddings), CORAL and WDN. Each color corresponds to a batch (i.e. domain), and there are ten batches in total. The negative controls appear in all the batches. The stopping time step is selected where the Silhouette score is maximized. WDN is better at aligning the distributions of the principal components across batches than the other methods.}
\label{fig:PC}
\end{figure}

\newpage

\newpage
\section{Conclusions}\label{sec:future_work}
We have shown how a neural network can be used to transform embedding vectors to `forget' specifically chosen domain information as indicated by our proposed domain classification accuracy metric. 
{The transformed embeddings still  preserve the underlying geometry of the space and maintain or even improve the k-NN MOA assignment metrics and the Silhouette score.}
Our approach uses the Wasserstein distance and can in principle handle fairly general distributions of embeddings (as long as the neural network used to approximate the Wasserstein function is general enough). Importantly, we do not have to assume that the distributions are Gaussian, which is required in much of existing literature.

The framework of our approach is quite general and extensible.
Unlike existing methods that are restricted to the negative controls for adjusting the embeddings, our method can also utilize information from replicates of multiple treatments across different domains.
However, the BBBC021 dataset used does not have treatment replicates across batches, so we have to rely on aligning based on the negative controls only. This means we implicitly assume that the transformations learned from the negative controls can be applied to all the other treatments. We expect our method to be more powerful in the context of experiments where many replicates are present, so that they can all be aligned simultaneously. We expect the transformations learned for such experiments to have better generalizability since it would use available knowledge from a greater portion of the embedding space.

Although methods using neural networks tend to be more flexible than traditional methods, they tend to be more difficult to train, in part due to hyperparameter tuning.
In the case of our method, we design a minimax problem that once optimized will remove the nuisance variation. However, we must use either early stopping or some form of regularization to prevent collapsing the embeddings together. 
Although 
\citet{shaham2018batch}
does not have the exact same problem in the  variational autoencoder setting, they instead need to either use a regularization parameter, or another hyperparameter to balance parts of the loss function associated with the removal of the batch effect and the preservation of the biological signal.
Meanwhile, \citet{amodio2018neuron} defined an explicit map in the latent space, which fixes the percentiles of the distributions to match. While this resolves the hyperparameter ambiguity, it also reduces the flexibility of the method and has to depend on certain assumptions of the latent space for the method to work.

We discuss potential future directions below, as well as other limiting issues.
One possible extension is to modify the form of the loss function~\label{eqn:our_form} by the following, which would more closely resemble finding the Wasserstein barycenter:
\begin{align}
\label{equ:new_form}
\sum_{i,j=1}^N W(\nu_i, A_{d_j}(\nu_j)).
\end{align}
The difference is that instead of comparing the pairwise transformed distributions, we compare the transformed distributions to the original distributions. One distinct advantage of this approach is that it avoids the `shrinking to a point' problem, and therefore does not require early stopping or a regularization term to converge to a meaningful solution. 
However, we have not found better performance for the new form of the loss function (eq.~\ref{equ:new_form}) for the BBBC021 dataset.


The Wasserstein functions were approximated by quite simple nonlinear functions, and it is possible that better results would be obtained using more sophisticated functions to capture the Wasserstein distance and its gradients more accurately. Similarly, The transformations $A_d$ could be generalized from affine to a more general class of nonlinear functions. As in \citet{shaham2017removal}, we expect ResNet would be natural candidates for these transformations.

We may fine-tune the Deep Metric Network used to generate the embeddings instead of training a separate network on top of its outputs (i.e. embeddings).
Another issue is how to weigh the various Wasserstein distances against each other. This might improve the results if there are many more points from some distributions than others (which happens quite often in real applications). 
Another extension may involve applying our method hierarchically to the various domains of the experiment. For example, we could apply our method on the plate level instead of the batch level only. 

Since the k-NN MOA assignment metrics and the Silhouette score are based on the cosine distance, it is possible that better results could be obtained by modifying the metric used to compute the Wasserstein distance accordingly, e.g. finding an optimal transportation plan only in non-radial directions.


\subsubsection*{Acknowledgments}
We would like to thank Mike Ando, Marc Coram, Marc Berndl, Subhashini Venugopalan, Arunachalam Narayanaswamy, Yaroslav Ganin, Luke Metz, Eric Christiansen, Philip Nelson, and Patrick Riley for useful discussions and suggestions.
\newpage

\newpage

\bibliography{bibliography}
\bibliographystyle{bib_style}

\newpage

\appendix

\begin{table}[!htbp]
\caption{
\textbf{NSC k-NN MOA assignment metrics ($\%$) for hand-engineered embeddings.} For all methods (Preprocessed, CORAL and WDN), we show the vanilla and bootstrap results. For WDN, we also show the results using two different stopping time steps – based on the average k-NN metric and the Silhouette score. The bootstrap results are represented by the mean and standard deviation (shown in parentheses) across the bootstrap estimates.}
\label{table:nsc}
\centering
\begin{tabular}{c|c|cccc}
\toprule
Method &  Type  &    1-NN &    2-NN &    3-NN &    4-NN \\
\midrule
\multirow{2}{*}{Preprocessed}    &  Vanilla     &  92.2 &  85.9 &  80.9 &  77.1 \\
\cline{2-6}
                        & Bootstrap    &  91.3 &  86.2 &  80.7 &  77.0 \\
                        &              &  (0.8) &  (0.6) &  (0.6) &  (0.6) \\
\midrule
\multirow{2}{*}{CORAL}  &  Vanilla     &  90.3 &  86.4 &  80.3 &  77.6 \\
\cline{2-6}
                        & Bootstrap    &  90.4 &  86.2 &  80.2 &  77.4 \\
                        &              &  (0.9) &  (0.7) &  (0.7) &  (0.6) \\
\midrule
\multirow{5}{*}{WDN}    & Max Silhouette  &  92.2 &  86.4 &  80.9 &  76.6 \\
\cline{2-6}
                        & Max Average k-NN  &  92.2 &  86.4 &  80.9 &  77.4 \\
\cline{2-6}
 & Bootstrap, max k-NN                 &  91.1 &  86.0 &  80.4 &  76.8 \\
                        &              &  (0.8) &  (0.6) &  (0.8) &  (0.7) \\
\cline{2-6}
 & Bootstrap, max Silhouette           &  91.1 &  86.0 &  79.8 &  76.4 \\
                        &              &  (1.0) &  (0.6) &  (0.7) &  (0.6) \\
\bottomrule
\end{tabular}
\end{table}

\begin{table}[!htbp]
\caption{\textbf{NSC NSB k-NN MOA assignment metrics ($\%$) for hand-engineered embeddings.} For all methods (Preprocessed, CORAL and WDN), we show the vanilla and bootstrap results. For WDN, we also show the results using two different stopping time steps – based on the average k-NN metric and the Silhouette score. The bootstrap results are represented by the mean and standard deviation (shown in parentheses) across the bootstrap estimates.}
\label{table:nsc-nsb}
\centering
\begin{tabular}{c|c|cccc}
\toprule
Method &  Type  &    1-NN &    2-NN &    3-NN &    4-NN \\
\midrule
\multirow{2}{*}{Preprocessed}    &  Vanilla     &  73.9 &  73.3 &  69.4 &  67.6 \\
\cline{2-6}
                        & Bootstrap    &  73.6 &  72.9 &  69.4 &  67.5 \\
                        &              &  (1.6) &  (1.3) &  (1.1) &  (0.9) \\
\midrule
\multirow{2}{*}{CORAL}  &  Vanilla     &  73.9 &  72.8 &  69.0 &  68.5 \\
\cline{2-6}
                        & Bootstrap    &  73.7 &  72.5 &  68.2 &  67.8 \\
                        &              &  (1.7) &  (1.2) &  (0.8) &  (0.8) \\
\midrule
\multirow{5}{*}{WDN}    & Max Silhouette  &  73.9 &  72.8 &  69.8 &  67.3 \\
\cline{2-6}
                        & Max Average k-NN  &  73.9 &  73.3 &  69.4 &  67.9 \\
\cline{2-6}
 & Bootstrap, max k-NN                 &  73.6 &  72.7 &  69.0 &  67.3 \\
                        &              &  (1.7) &  (1.3) &  (1.2) &  (1.0) \\
\cline{2-6}
 & Bootstrap, max Silhouette           &  73.5 &  72.3 &  68.2 &  67.0 \\
                        &              &  (1.6) &  (1.2) &  (0.9) &  (0.8) \\
\bottomrule
\end{tabular}
\end{table}

\begin{table}[!htbp]
\centering
\caption{\textbf{Silhouette scores and batch classification accuracy for hand-engineered embeddings.} We compare the methods Preprocessed, CORAL and WDN. The Silhouette score is evaluated over the entire dataset (excluding the negative control), and the batch classification accuracy is evaluated over the negative control, both at the time step where the Silhouette score is maximized. The bootstrap results are represented by the mean and standard deviation (denoted by $\sigma(\cdot)$) across the bootstrap estimates.}
\label{table:sil}
\resizebox{\columnwidth}{!}{%
\begin{tabular}{lrlrlrl}
\toprule
{} &    WDN & $\sigma$\text{(WDN)} &  CORAL & $\sigma$\text{(CORAL)} &    Preproc. & $\sigma$\text{(Preproc.)} \\
\midrule
Silhouette Score              &  0.438 &                       &  0.440 &                         &  0.436 &                       \\
Bootstrap Silhouette Score    &  0.429 &               0.003 &  0.433 &                 0.003 &  0.429 &               0.003 \\
Bootstrap Logistic Regression &  30.0\% &                 0.8\% &  33.3\% &                   0.9\% &  33.9\% &                 0.9\% \\
Bootstrap Random Forest       &  27.3\% &                 0.2\% &  27.2\% &                   0.2\% &  29.5\% &                 0.8\% \\
\bottomrule
\end{tabular}
}
\end{table}

\begin{table}[!htbp]
\centering
\caption{
\textbf{NSC k-NN MOA assignment metrics ($\%$) for DNN embeddings.} For all methods (TVN, CORAL and WDN), we show the vanilla and bootstrap results. For WDN, we also show the results using two different stopping time steps – based on the average k-NN metric and the Silhouette score. The bootstrap results are represented by the mean and standard deviation (shown in parentheses) across the bootstrap estimates.}
\label{table:deep-nsc}
\begin{tabular}{c|c|cccc}
\toprule
Method &  Type  &    1-NN &    2-NN &    3-NN &    4-NN \\
\midrule
\multirow{2}{*}{TVN}    &  Vanilla     &  96.1 &  92.2 &  89.3 &  87.4 \\
\cline{2-6}
                        & Bootstrap    &  94.7 &  92.1 &  89.1 &  87.4 \\
                        &              &  (1.0) &  (0.6) &  (0.6) &  (0.5) \\
\midrule
\multirow{2}{*}{CORAL}  &  Vanilla     &  97.1 &  94.7 &  90.9 &  88.4 \\
\cline{2-6}
                        & Bootstrap    &  96.8 &  94.5 &  91.1 &  89.0 \\
                        &              &  (0.5) &  (0.5) &  (0.6) &  (0.5) \\
\midrule
\multirow{5}{*}{WDN}    & Max Silhouette  &  97.1 &  95.1 &  92.6 &  90.2 \\
\cline{2-6}
                        & Max Average k-NN  &  97.1 &  95.1 &  92.6 &  90.2 \\
\cline{2-6}
 & Bootstrap, max k-NN                 &  96.4 &  94.2 &  91.5 &  89.7 \\
                        &              &  (0.8) &  (0.8) &  (0.7) &  (0.6) \\
\cline{2-6}
 & Bootstrap, max Silhouette           &  96.5 &  94.3 &  91.8 &  89.8 \\
                        &              &  (0.7) &  (0.7) &  (0.7) &  (0.6) \\
\bottomrule
\end{tabular}
\end{table}

\begin{table}[!htbp]
\centering
\caption{\textbf{NSC NSB k-NN MOA assignment metrics ($\%$) for DNN embeddings.} For all methods (TVN, CORAL and WDN), we show the vanilla and bootstrap results. For WDN, we also show the results using two different stopping time steps – based on the average k-NN metric and the Silhouette score. The bootstrap results are represented by the mean and standard deviation (shown in parentheses) across the bootstrap estimates.}
\label{table:deep-nsc-nsb}
\begin{tabular}{c|c|cccc}
\toprule
Method &  Type  &    1-NN &    2-NN &    3-NN &    4-NN \\
\midrule
\multirow{2}{*}{TVN}    &  Vanilla     &  90.2 &  88.9 &  84.7 &  83.9 \\
\cline{2-6}
                        & Bootstrap    &  90.2 &  88.2 &  84.0 &  83.4 \\
                        &              &  (1.1) &  (0.9) &  (0.6) &  (0.6) \\
\midrule
\multirow{2}{*}{CORAL}  &  Vanilla     &  91.3 &  89.4 &  85.1 &  84.5 \\
\cline{2-6}
                        & Bootstrap    &  92.5 &  89.4 &  84.9 &  84.5 \\
                        &              &  (1.2) &  (0.8) &  (0.6) &  (0.5) \\
\midrule
\multirow{5}{*}{WDN}    & Max Silhouette  &  93.5 &  91.7 &  85.8 &  84.8 \\
\cline{2-6}
                        & Max Average k-NN  &  93.5 &  91.7 &  85.8 &  84.8 \\
\cline{2-6}
 & Bootstrap, max k-NN                 &  92.2 &  90.3 &  85.2 &  84.5 \\
                        &              &  (1.2) &  (1.0) &  (0.7) &  (0.6) \\
\cline{2-6}
 & Bootstrap, max Silhouette           &  92.6 &  90.8 &  85.8 &  84.8 \\
                        &              &  (0.9) &  (0.8) &  (0.7) &  (0.5) \\
\bottomrule
\end{tabular}
\end{table}

\begin{table}[!htbp]
\centering
\caption{\textbf{Silhouette scores and batch classification accuracy for DNN embeddings.} We compare the methods TVN, CORAL and WDN. The Silhouette score is evaluated over the entire dataset (excluding the negative control), and the batch classification accuracy is evaluated over the negative control, both at the time step where the Silhouette score is maximized. The bootstrap results are represented by the mean and standard deviation (denoted by $\sigma(\cdot)$) across the bootstrap estimates.}
\label{table:deep-sil}
\begin{tabular}{lrlrlrl}
\toprule
{} &    WDN & $\sigma$\text{(WDN)} &  CORAL & $\sigma$\text{(CORAL)} &    TVN & $\sigma$\text{(TVN)} \\
\midrule
Silhouette Score              &  0.514 &                       &  0.510 &                         &  0.504 &                       \\
Bootstrap Silhouette Score    &  0.505 &               0.002 &  0.502 &                 0.002 &  0.496 &               0.002 \\
Bootstrap Logistic Regression &  43.2\% &                 0.6\% &  66.4\% &                   0.7\% &  63.6\% &                  1.0\% \\
Bootstrap Random Forest       &  34.6\% &                 1.1\% &  48.6\% &                   0.6\% &  45.9\% &                 0.2\% \\
\bottomrule
\end{tabular}
\end{table}

\end{document}